\documentclass[10pt,twocolumn,letterpaper]{article}
\usepackage[margin=0.75in,top=0.72in,bottom=0.73in,columnsep=0.24in]{geometry}
\usepackage[T1]{fontenc}
\usepackage[utf8]{inputenc}
\usepackage{times}
\usepackage{graphicx}
\usepackage{amsmath,amssymb}
\usepackage{booktabs,multirow}
\usepackage{cite}
\usepackage{url}
\usepackage[hidelinks]{hyperref}
\usepackage{microtype}
\usepackage{balance}
\title{Does It Depend on the Split? Frozen Satellite Features vs. Fine-Tuned CNNs\\for Tree Species Mapping}
\author{Dharani Deivasihamani\qquad Nandakishore Puttashamachar\\[0.4em]
SatSure Analytics India Pvt. Ltd.\\
\texttt{\{dharani, nandakishore\}@satsure.co}}
\date{}
\begin{document}
\maketitle
\thispagestyle{plain}
\begin{abstract}
Self-supervised foundation models are often said to need fewer labels than supervised training. We test this on a forestry task: classifying 20 tree species classes in 0.2 m aerial images from the TreeSatAI dataset. We compare a frozen DINOv3 model pretrained on 493 million satellite images, with only a linear classifier trained on top, against an ImageNet ResNet-50 that is fine-tuned end to end. On the official split, fine-tuning wins clearly (macro F1 0.764 vs. 0.685). However, 28.8\% of the official test patches overlap a training or validation patch, and copying the label of the nearest such patch is correct for 83.6\% of them. We therefore build a spatial split that holds out whole 2 km squares with a 200 m buffer. On this split the comparison becomes inconclusive (0.648 vs. 0.640, paired 95\% CI of the difference [\ensuremath{-}0.054, 0.016]), and the learning curves of the two models overlap at every label budget from 343 to 33,126 training patches. On the five rarest classes, the frozen model achieves a higher mean F1 (0.566 vs. 0.445), although the confidence interval for this difference includes zero ([\ensuremath{-}0.05, 0.23]). Diagnostics show that the frozen features mainly miss differences between closely related species, such as the two native oaks. Choosing the input size and layer on validation data (224 px, block 16) raises the frozen model to 0.662 on the spatial split. A comparison that looks decisive on the official split is thus inconclusive under geographically separated evaluation. We release the spatial split and our code.
\end{abstract}
\section{Introduction}
Forest managers need to know which tree species grow where. Species maps are used to plan harvesting and replanting, to find stands at risk from drought or pests \cite{ref10}, and, together with canopy analysis, to support agroforestry and the monitoring of agricultural and forest land. Many agencies fly regular aerial surveys, so recognising species from these images is of clear practical value. The hard cases are the rare species. They have few labels, and they are often the ones that matter most for biodiversity and for planning. Self-supervised foundation models are a promising answer to this problem. DINOv3 \cite{ref28} includes a vision transformer pretrained without labels on SAT-493M, a set of 493 million satellite images. If its features already separate tree species, a user can run the model once over an image archive and train only a linear classifier on top. This is cheap, and it should need fewer labels. Recent benchmarks, however, find that geospatial foundation models do not consistently beat supervised models \cite{ref21}, and a well-tuned ImageNet model remains a strong baseline \cite{ref8}. We first ran this comparison on the official split of TreeSatAI \cite{ref1}. A fine-tuned ImageNet ResNet-50 beat the frozen DINOv3 model by 0.08 macro F1, also on the rare classes. We then looked at where the test patches are. Forest labels are spatially clustered, and a model can score well by recognising the place rather than the species \cite{ref14,ref25}. In the official split, 28.8\% of the test patches overlap a training or validation patch, and the label of the nearest such patch alone is correct for 83.6\% of them. Such a test rewards a model that adapts closely to the training stands. It does not tell us how well a model works in a new area, which is how species maps are used in practice. We therefore repeat the comparison on a spatial split that holds out whole 2 km squares (Fig. 1). On this split, both models lose accuracy, the fine-tuned model more so, and the difference between them is no longer significant. The largest change is on the rarest species. Here the fine-tuned model loses 0.27 F1 and the frozen model only 0.10, although the remaining difference between them is uncertain. This fits earlier findings that fine-tuning can distort pretrained features and hurt out-of-distribution performance \cite{ref16}, and that self-supervised features are less affected by class imbalance \cite{ref20}. Our design compares two complete strategies, a frozen satellite-pretrained transformer and a fine-tuned ImageNet CNN. It cannot separate the effects of architecture, pretraining data and freezing, so our claims are about the comparison of these strategies and how it depends on the evaluation split, not about self-supervised pretraining being better in general. Our main contributions are:
\begin{itemize}
\item We show that the official TreeSatAI split overlaps spatially with its training data, and that test patches close to training data favour the fine-tuned model more than the frozen one (Sec. 3.1).
\item We release a spatial split with 2 km squares and a 200 m buffer. On it, a frozen DINOv3 model with a linear classifier performs on par with a fine-tuned ResNet-50 at every label budget we tested (Sec. 5).
\item On unseen areas, the frozen model reaches a higher mean F1 on the five rarest species, but the confidence interval of the difference includes zero because these classes have few test locations (Sec. 5.1).
\item We locate the remaining errors of the frozen features in closely related species. Colour, classifier capacity and pooling do not explain them. A slightly coarser input and a middle layer, chosen on validation data, raise the frozen model to 0.662 macro F1 on the spatial test set (Sec. 6).
\end{itemize}

\begin{figure*}[!t]
\centering
\includegraphics[width=\textwidth]{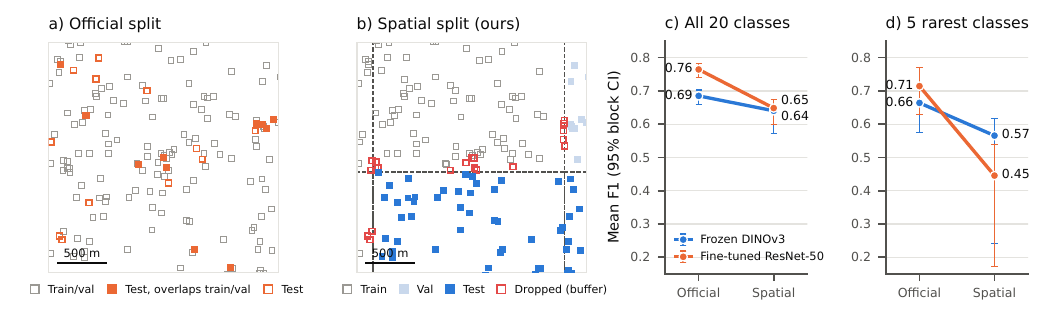}
\caption{Official vs. spatial evaluation on TreeSatAI. (a) In the official split, test patches lie between training patches, and many overlap one (filled squares). The same 2.4 km window is shown in both maps; each square is one 60.8 m patch. (b) Our split assigns whole 2 km squares to training, validation or test, and drops patches within 200 m of a square with another role. (c) Over all 20 classes, the clear lead of the fine-tuned ResNet-50 on the official split becomes inconclusive on the spatial split. (d) On the five rarest classes, the fine-tuned model loses much more than the frozen DINOv3 model; the remaining difference on the spatial split has a confidence interval that includes zero. Error bars are 95\% confidence intervals from resampling whole 2 km squares.}
\label{fig:1}
\end{figure*}
\section{Related Work}
\textbf{Tree species from remote sensing.} Fassnacht et al. \cite{ref10} reviewed decades of tree species classification from remote sensing; convolutional networks on drone images map species well within a site \cite{ref26}. TreeSatAI \cite{ref1} combines 0.2 m aerial images with Sentinel-1 and Sentinel-2 data and with labels from forest inventories in Lower Saxony, Germany. Its benchmarks are multi-label and per genus, and thus not directly comparable with our task. On the aerial images, a frozen ImageNet ResNet-18 with a retrained last layer reached only about half the F1 of a fine-tuned one \cite{ref1}. Their test set is a fixed 10\% of the data; its spatial layout is not discussed. PureForest \cite{ref11} provides 0.2 m aerial images and lidar for 13 classes of single-species forests in France. It assigns whole forests to training, validation or test, without a buffer, and a ResNet-18 on the aerial images reached 73.1\% overall accuracy.

\textbf{Foundation models for Earth observation.} Self-supervised vision transformers \cite{ref9} such as DINO \cite{ref6} and DINOv2 \cite{ref23} learn general image features. These features are usually tested with a linear classifier, a linear probe \cite{ref2}. Several models have been pretrained on satellite data, for example with masked autoencoding \cite{ref7}, and DINOv2 features trained on satellite images have been used to map canopy height \cite{ref29}. DINOv3 \cite{ref28} provides a ViT-L/16 pretrained on SAT-493M, which we use. GEO-Bench \cite{ref18} and PANGAEA \cite{ref21} compare such models across tasks; PANGAEA finds that they do not consistently outperform supervised models. Corley et al. \cite{ref8} show that resizing and normalisation choices can change these comparisons a lot. OmniSat \cite{ref3} and AnySat \cite{ref4} report multi-label TreeSatAI results. For species, Ishikawa et al. \cite{ref13} find that pretrained Sentinel-based embeddings beat hand-crafted features on Dutch forest inventory plots.

\textbf{Frozen versus fine-tuned features under shift.} Fine-tuning all weights usually gives the best in-distribution accuracy. Kumar et al. \cite{ref16} show, however, that it can distort pretrained features and do worse than a linear probe out of distribution. Liu et al. \cite{ref20} find that self-supervised features suffer less from class imbalance than supervised ones. They suggest that self-supervised training also learns features from the frequent classes that transfer to the rare ones. WILDS \cite{ref15} collects benchmarks in which models are tested on new locations or domains. Our study tests these ideas on a forestry task in which both the shift (a new area) and the imbalance (rare species) are natural.

\textbf{Spatial validation.} When nearby samples are similar, random splits give overly optimistic results \cite{ref22,ref24,ref25}. Kattenborn et al. \cite{ref14} showed this for deep networks on high-resolution vegetation images. A recent study of foundation models for satellite time series reported a significant accuracy drop for tree species in regions not seen during training \cite{ref5}. The usual remedy is to split the data into spatial blocks \cite{ref25,ref30} and to compute confidence intervals by resampling whole blocks \cite{ref17}. We follow both practices and also measure how the gap between two kinds of models changes.
\section{Data and Splits}
\textbf{TreeSatAI.} We use the aerial images of the TreeSatAI Benchmark Archive v1.0.1 \cite{ref1,ref27} (CC-BY-4.0). It contains 50,381 patches of 304 \ensuremath{\times} 304 pixels at 0.2 m, so each patch covers 60.8 \ensuremath{\times} 60.8 m. Each patch has one of 20 labels: 19 tree species or genera, and Cleared (clear-cut land; Fig. 2). The classes are unbalanced. The most common, Pinus sylvestris, has 13.1\% of the patches, and the five rarest (Tilia, Prunus, Pinus nigra, Populus and Pinus strobus) have 143--318 training patches each. The images have four bands stored in the order near-infrared, green, blue, and red, although the file metadata suggests standard RGB. We build true-colour images from bands 4, 2, and 3 and check this with an automated test. Both models see only these three bands.
\subsection{The official split and its overlap}
The official split has 45,337 training and 5,044 test patches. We hold out 20\% of the training patches, stratified by class, for validation. Labels in TreeSatAI are strongly clustered in space. Two patches less than 61 m apart have the same label 85\% of the time, compared with 8\% for two random patches. In the official split, 28.8\% of the test patches overlap a training or validation patch (centre distance below 60.8 m), and 84.6\% have one within 200 m. To see how much this matters, we use a trivial predictor that ignores the image and copies the label of the nearest training or validation patch. On the 1,453 overlapping test patches it is correct 83.6\% of the time (83.2\% when only training patches are used), more often than either model in this study. We then compare recall within each class at different distances to the nearest training or validation patch, so that the different class mix at each distance does not bias the result (Fig. 3). Both models do better on test patches close to training data, but the fine-tuned ResNet-50 gains more. Its recall falls by 0.12 between overlapping and distant patches (0.816 to 0.694), while the frozen DINOv3 model falls by 0.07 (0.732 to 0.658). This suggests that the official test rewards a model for adapting to the training stands, and that fine-tuning does this more than a linear probe on frozen features.
\subsection{Our spatial split}
TreeSatAI has no stand identifiers, so we group patches into 2 \ensuremath{\times} 2 km squares by their centres (EPSG:25832 coordinates). The squares keep geographically neighbouring patches from the same local landscape together, so that they are not spread across training and evaluation sets. Because patches in adjacent squares can still be close, we also apply a buffer, described below. We assign whole squares to test (one seventh) and then to validation (one fifth of the rest), stratified by class (scikit-learn StratifiedGroupKFold, seed 42). We then drop every training or validation patch whose centre is within 200 m of a test patch, and every training patch within 200 m of a validation patch. This removes 1,530 patches. The final split has 33,126 training, 8,493 validation and 7,232 test patches in 1,754, 430 and 355 squares, and every class occurs in each part (Fig. 1b). We chose 2 km blocks and a 200 m buffer before test evaluation to limit proximity across splits while retaining rare-class locations. Larger buffers were not tested. We release the split file with its SHA-256 checksum, a fingerprint of the file that lets others check that their copy is identical to ours.
\begin{figure*}[!t]
\centering
\includegraphics[width=\textwidth]{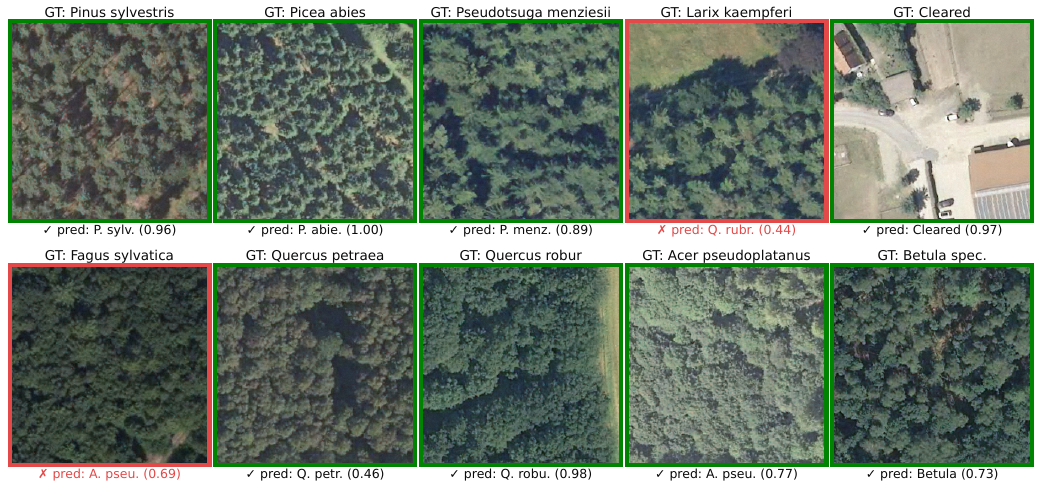}
\caption{Randomly drawn (not hand-picked) patches from the spatial test set, one for each of ten classes. Each 60.8 m patch has one label (GT). Below each patch is the prediction of the frozen DINOv3 model with its confidence; a green frame and $\checkmark$ mark a correct prediction, a red frame and $\times$ a wrong one.}
\label{fig:2}
\end{figure*}
\begin{figure}[!t]
\centering
\includegraphics[width=\columnwidth]{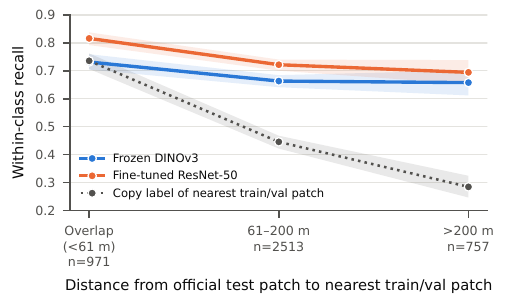}
\caption{Official test set: recall within each class against the distance to the nearest training or validation patch. To avoid a different class mix in each group, we keep only the 11 classes with at least 10 test patches in every group, so the group sizes (n, 4,241 patches in total) are smaller than the full test set (5,044 patches, of which 1,453 overlap). The dotted line copies the label of that nearest patch and does not look at the image. Bands are 95\% intervals from resampling 2 km squares.}
\label{fig:3}
\end{figure}
\section{Models and Protocol}
\textbf{Frozen DINOv3 with a linear probe.} We use the DINOv3 ViT-L/16 pretrained on SAT-493M \cite{ref28} (303M parameters, none updated), loaded from the official code and verified by checksum. Each 304 \ensuremath{\times} 304 patch is used at full size, which gives a 19$\times$19 grid of tokens, and is normalised with the SAT-493M statistics. We store the last-layer CLS token and the mean of the patch tokens (1024 dimensions each) for every patch. This is done once. On these vectors we train a multinomial logistic regression with an L2 penalty, using L-BFGS \cite{ref19} on standardised features. We choose the feature type (CLS token, token mean or both), the class weighting and the penalty on validation macro F1. Validation selected CLS and token mean together on the official split and the token mean alone on the spatial split, both with no class weighting and \ensuremath{\lambda} = 10$^{-3}$. Training takes under three minutes and less than 1 GB of GPU memory.

\textbf{Fine-tuned ResNet-50.} As the supervised reference we fine-tune all weights of an ImageNet-pretrained ResNet-50 \cite{ref12} on the same training patches at full size. We use AdamW (weight decay 0.05), batch size 128, one warm-up epoch, label smoothing 0.1, random flips and 90$^{\circ}$ rotations, and up to 30 epochs with early stopping (patience 8). We try learning rates 10$^{-4}$ and 3·10$^{-4}$ and keep the epoch with the best validation macro F1. Neither model uses class weighting, so the rare-class comparison is not affected by different loss weights.

\textbf{Hardware and cost.} All runs use one NVIDIA L40S GPU (48 GB) with bfloat16 arithmetic. Table 1 lists the measured costs. Training the frozen pipeline needs one pass of the backbone over the images and a few minutes for the linear classifier, all in less than 3 GB of GPU memory. Fine-tuning ResNet-50 takes about an hour and 11 GB per split. At inference the ViT-L backbone is slower than ResNet-50, as expected for a model with about 13 times more parameters (303M vs. 23.5M).

\textbf{Metrics and test protocol.} Our main metric is macro F1, so that each class counts equally. We also report balanced accuracy, accuracy and top-3 accuracy. We compute 95\% confidence intervals by resampling whole 2 km squares 2,000 times \cite{ref17}; classes absent from a resample are left out of its mean. For differences between the two models we resample the same squares for both (paired). All choices are made on validation data, and each test set is evaluated once per model. All later analyses use the stored test predictions.
\begin{table}[!t]
\caption{Measured cost on one NVIDIA L40S (bfloat16, batch 128 unless noted; throughput excludes reading images from disk). Training times are for the full search on one split.}
\centering\scriptsize
\resizebox{\columnwidth}{!}{\begin{tabular}{lrr}\toprule
Step & Time & Peak GPU mem.\\\midrule
DINOv3 features, 304 px (all 50,381 patches) & 3.4 min (247 img/s) & 3.0 GB\\
DINOv3 features, 224 px, block 16 & 1.8 min (462 img/s) & 2.4 GB\\
Linear classifier, full $\lambda$ search & $<3$ min & $<1$ GB\\
ResNet-50 fine-tuning, LR search & 63--76 min & 11.0 GB\\\midrule
Inference, DINOv3 304 px (batch 1) & 4.0 ms/img (12.8 ms) & 3.0 GB\\
Inference, ResNet-50 304 px (batch 1) & 0.5 ms/img (3.4 ms) & 1.4 GB\\\bottomrule
\end{tabular}}
\label{tab:cost}
\end{table}
\begin{table}[!t]
\caption{Test results on both splits. Brackets: 95\% CI from resampling 2 km squares. $\Delta$: DINOv3 (304 px, last block) minus ResNet-50 macro F1 with a paired CI. The last DINOv3 row uses the input size and block chosen on spatial validation data (Sec. 6). The two splits use different test sets.}
\centering\scriptsize
\resizebox{\columnwidth}{!}{\begin{tabular}{llrrr}\toprule
Split & Model & Macro F1 & Bal. acc. & Top-3\\\midrule
\multirow{4}{*}{Official $n$=5,044} & ResNet-50, fine-tuned & 0.764 [0.740, 0.782] & 0.747 & 0.928\\
& DINOv3, frozen + linear & 0.685 [0.660, 0.703] & 0.668 & 0.898\\
& $\Delta$ & $-0.079$ [$-0.099$, $-0.061$] & &\\
& DINOv3, 224 px, block 16 & 0.696 [0.670, 0.715] & 0.676 & 0.907\\\midrule
\multirow{4}{*}{Spatial $n$=7,232} & ResNet-50, fine-tuned & 0.648 [0.598, 0.674] & 0.633 & 0.890\\
& DINOv3, frozen + linear & 0.640 [0.572, 0.656] & 0.617 & 0.883\\
& $\Delta$ & $-0.008$ [$-0.054$, 0.016] & &\\
& DINOv3, 224 px, block 16 & 0.662 [0.587, 0.677] & 0.641 & 0.893\\\bottomrule
\end{tabular}}
\label{tab:results}
\end{table}
\begin{figure}[!t]
\centering
\includegraphics[width=\columnwidth]{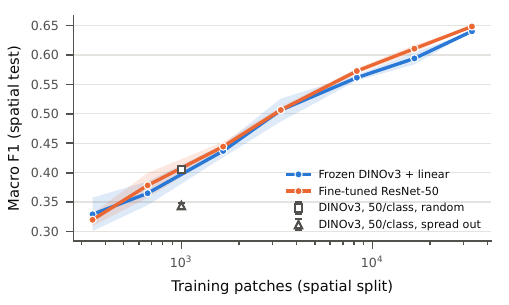}
\caption{Spatial test macro F1 against the number of training patches (mean $\pm$ s.d. over 3 random class-stratified subsets; one run at 100\%). Grey markers: DINOv3 trained on 50 patches per class, drawn at random or spread as far apart as possible.}
\label{fig:4}
\end{figure}
\begin{figure}[!t]
\centering
\includegraphics[width=\columnwidth]{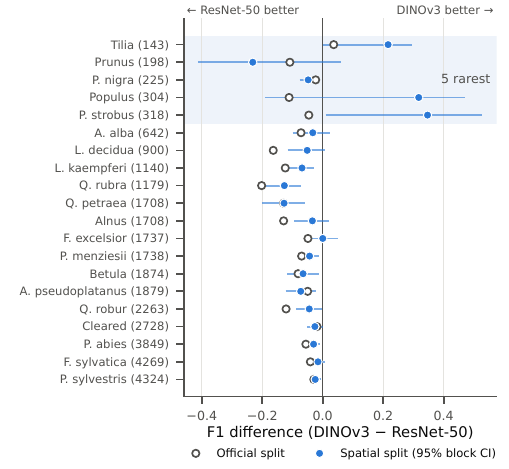}
\caption{F1 difference between frozen DINOv3 and fine-tuned ResNet-50 for each class, ordered by the number of training patches (in brackets). Open circles: official split. Filled: spatial split, with 95\% intervals from resampling 2 km squares. Shaded: the five rarest classes.}
\label{fig:5}
\end{figure}
\section{Results}
\textbf{Main comparison.} On the official split, the fine-tuned ResNet-50 is better by 0.079 macro F1, and the paired interval excludes zero (Tab. 2). On the spatial split, ResNet-50 loses 0.116 and DINOv3 only 0.045; the remaining gap of 0.008 has an interval that includes zero. Colour statistics alone reach only 0.18--0.19. The change is not limited to a few classes. For 16 of the 20 classes, the per-class difference moves in favour of DINOv3 from the official to the spatial split (Fig. 5).

\textbf{Label efficiency.} A common argument for frozen foundation features is that they need fewer labels. We trained both models on class-stratified subsets of 1\% to 100\% of the spatial training set, three subsets per size, and tested each once on the spatial test set (Fig. 4). The two curves overlap at every size. With 343 training patches (1\%), DINOv3 reaches 0.329 and ResNet-50 0.320; with 3,312 patches (10\%), both reach 0.506. We find no budget at which the frozen model is clearly more label-efficient. With 50 patches per class, choosing them as far apart as possible gave a lower score (0.344) than drawing them at random (0.406), probably because isolated patches are often unusual stands.

\subsection{Rare classes in new areas}
The rarest classes show the clearest difference between the two splits (Fig. 1d, Fig. 5). On the official split, ResNet-50 is better on the five rarest classes as well, with a mean F1 of 0.714 against 0.664 (paired difference \ensuremath{-}0.051, CI [\ensuremath{-}0.107, 0.005]). On the spatial split the order reverses: DINOv3 reaches 0.566 and ResNet-50 0.445. The fine-tuned model loses 0.27 F1 on these classes, the frozen model 0.10. The paired difference is +0.12 in favour of DINOv3, with a CI of [\ensuremath{-}0.05, 0.23]; DINOv3 is ahead in 89\% of the resamples. The gain comes from three classes. For Pinus strobus the difference is +0.35 (CI [0.01, 0.53]), for Tilia +0.22 ([0.00, 0.30]) and for Populus +0.32 ([\ensuremath{-}0.19, 0.47]). In all three, ResNet-50 rarely predicts the class in new areas. Its recall is 0.13 for P. strobus, 0.27 for Tilia and 0.11 for Populus, against 0.40, 0.50 and 0.35 for DINOv3. On the official split, ResNet-50 recalled the same three classes as well as or better than DINOv3 (0.27--0.89 against 0.32--0.84). Two rare classes do not follow this pattern. ResNet-50 stays better on Prunus (\ensuremath{-}0.23, CI [\ensuremath{-}0.41, 0.06]) and slightly better on Pinus nigra (\ensuremath{-}0.05), which both models recognise well. These results indicate, but do not establish, that the frozen strategy transfers better to new areas for species with few labels. Because the two models also differ in architecture and pretraining data, this result describes the two strategies, not self-supervised pretraining as such. A fine-tuned network can fit the few training stands of a rare species closely, which helps when test patches come from those same stands and hurts when they do not. This agrees with earlier work on fine-tuning under shift \cite{ref16} and on self-supervised features under class imbalance \cite{ref20}. The uncertainty is large, however, because the rare classes come from few places. All 110 spatial test patches of P. nigra lie in one 2 km square, and 73\% of the Tilia and 77\% of the Populus test patches each lie in one square.
\begin{table}[!t]
\caption{Diagnostics on the spatial validation set (macro F1). Top: feature and classifier variants at full resolution. Bottom: token mean from transformer block 16 or 24 (of 24) after resizing each patch; GSD is the resulting ground sampling distance.}
\centering\scriptsize
\begin{tabular}{lr}\toprule
Variant (304 px, last block) & Val. macro F1\\\midrule
Colour statistics only, linear & 0.153\\
CLS token, linear & 0.522\\
Token mean, linear (used) & 0.576\\
Token mean + colour statistics, linear & 0.578\\
Token mean, 2-layer MLP & 0.564\\
Token mean, linear, on its training set & 0.862\\\bottomrule
\end{tabular}
\medskip
\begin{tabular}{lrrr}\toprule
Input size & GSD & Block 16 & Block 24\\\midrule
96 px & 0.63 m & 0.534 & 0.529\\
160 px & 0.38 m & 0.580 & 0.561\\
224 px & 0.27 m & 0.603 & 0.579\\
304 px & 0.20 m & 0.591 & 0.578\\\bottomrule
\end{tabular}
\label{tab:diagnostics}
\end{table}
\section{Where Frozen Features Fall Short}
On the spatial split, the two models are level overall, but they make different errors. There are 816 test patches that only ResNet-50 classifies correctly and 552 that only DINOv3 classifies correctly. Most of the 816 are confusions between closely related or similar-looking broadleaf species: the two native oaks Q. petraea and Q. robur (80 patches), Acer and Fagus (46), Acer and Fraxinus (38), and Alnus and Betula (36). At a coarse level (conifer, broadleaf or cleared) the two models are close, at 0.904 and 0.924 accuracy. To find out why the frozen features miss these fine differences, we ran the diagnostics in Tab. 3. They use only the spatial training and validation sets; the test set was not used. \textbf{Not colour, not the classifier.} Adding colour statistics changes nothing (0.576 to 0.578), a two-layer MLP does not beat the linear classifier (0.564), and the linear classifier fits its training set well (0.862). The limit is in what the features generalise to new areas, not in colour or classifier capacity.

\textbf{Resolution and depth.} SAT-493M was pretrained on 0.6 m imagery, while TreeSatAI patches have 0.2 m pixels, so trees look about three times larger than during pretraining. Shrinking the patches to the pretraining scale (96 px, 0.63 m) is the worst choice, because it removes the crown texture that separates species. A mild reduction to 224 px (0.27 m) works best. Middle blocks also help: at every input size, block 16 of the 24 transformer blocks beats the last block, and the best setting, 224 px with block 16, reaches 0.603 against 0.576. Middle layers probably keep more local texture, which would help with fine-grained classes. We then tested this validation-selected setting once on both test sets (Tab. 2). It raises spatial macro F1 from 0.640 to 0.662 (paired gain +0.021, CI [\ensuremath{-}0.001, 0.037]), which puts the frozen model slightly ahead of the fine-tuned ResNet-50 (+0.013, CI [\ensuremath{-}0.042, 0.038]). On the five rarest classes it reaches 0.602 against 0.445 for ResNet-50 (+0.16, CI [\ensuremath{-}0.04, 0.26]). On the official split it gains 0.011 (0.696). Input scale and layer choice should therefore be part of the validation search when frozen satellite models are applied to aerial images \cite{ref8}. Because an earlier version of these diagnostics swapped the labels of blocks 16 and 24, 224 px with the last block was also tested once (0.660 spatial).
\section{Discussion and Limitations}
\textbf{What this means in practice.} On new areas, a frozen DINOv3 SAT-493M model with a linear classifier performed on par with a fine-tuned ResNet-50, at every training-set size we tried, and it reached a higher mean F1 on the five rarest species, although that difference is not significant. It is also much cheaper to train (Tab. 1). Features are extracted once, in a few minutes, and each new classifier then trains in under three minutes with less than 1 GB of GPU memory, compared with about an hour and 11 GB for every fine-tuning run. This matters when a model must be retrained often, for new regions, label sets or survey years. The price is slower inference, so for very large archives the one-time feature extraction is the main cost. At macro F1 near 0.65, neither model should guide decisions without validation. The official split would have led to the opposite conclusion. Spatial splitting evaluates generalization, not accuracy improvement; we recommend spatial tests with block-level confidence intervals. Comparing the two model strategies is sensitive to overlap because they adapt differently to nearby training stands. \textbf{Relation to earlier DINOv3 results.} Our results do not contradict the reported strength of DINOv3 features \cite{ref28}. The frozen model was never trained on this task, this sensor or this resolution: SAT-493M contains 0.6 m satellite images, while TreeSatAI uses 0.2 m aerial images. With only a linear layer on top, it still performs on par with a network whose every weight was tuned on the target data, and it does so on new areas, where species maps are actually used. The lead of fine-tuning on the official split mainly reflects how well a model adapts to the training stands. The remaining errors are between closely related species, and a better choice of input size and layer already reduces them, which suggests that a linear probe on the last layer does not use all the information in the features.

\textbf{Limitations.} (i) Few locations for rare classes. The rare-class result rests on a few spatial blocks. It points in the same direction for three of five classes, but its overall interval includes zero. (ii) One dataset and one reference model. We used one region, one CNN and one foundation model, so the comparison mixes architecture, pretraining data and freezing strategy and cannot attribute differences to any one of them. We did not fine-tune DINOv3, which could combine the strengths of both. (iii) Data. We use 2 km squares instead of stands (one split, buffer and seed), only RGB bands (near-infrared and Sentinel data would likely help the broadleaf species), and inventory labels with some noise.
\section{Conclusion}
On TreeSatAI, the official split makes a fine-tuned ResNet-50 look clearly better than a frozen DINOv3 model. More than a quarter of its test patches overlap the training or validation data, and test patches near training data favour the fine-tuned model most. On a split that holds out whole 2 km squares, the difference is no longer significant, and the learning curves of the two models overlap at every label budget. The frozen DINOv3 features generalise well to new areas: moving from the official to the spatial split, they lose much less than the fine-tuned network, both over all classes (0.045 vs. 0.116 macro F1) and on the five rarest classes (0.10 vs. 0.27). On these rare classes the frozen model also reaches a higher mean F1 (0.566 vs. 0.445, and 0.602 with the validation-selected input size and layer), although the confidence interval of the difference still includes zero. Without any training of the backbone, self-supervised satellite features thus hold up well exactly where species maps are needed: in unseen areas and for species with few labels. A model comparison that looks decisive under the official split can thus become inconclusive under geographically separated evaluation, and tree species models should be compared on new areas. Next, we will test whether this holds on other forest datasets, such as PureForest \cite{ref11}, and with a fine-tuned DINOv3.
\textbf{Reproducibility.} Code, configuration files, the spatial split (with SHA-256 checksums to verify the files), stored test predictions and evaluation logs are available at huggingface.co/Dharani1207/treesatai-spatial-dinov3. Python 3.12, PyTorch 2.8, DINOv3 commit 6876159a.

\textbf{Acknowledgements.} We thank the authors of TreeSatAI for releasing the dataset under an open licence, and Meta AI for releasing the DINOv3 code and weights. Computing resources were provided by SatSure Analytics India Pvt. Ltd.

\balance


\begin{thebibliography}{30}
\bibitem{ref1} Steve Ahlswede, Christian Schulz, Christiano Gava, Patrick Helber, Benjamin Bischke, Michael F{o}rster, Florencia Arias, J{o}rn Hees, Beg{u}m Demir, and Birgit Kleinschmit. TreeSatAI benchmark archive: a multi-sensor, multi-label dataset for tree species classification in remote sensing. Earth System Science Data, 15(2):681--695, 2023.

\bibitem{ref2} Guillaume Alain and Yoshua Bengio. Understanding intermediate layers using linear classifier probes. arXiv preprint arXiv:1610.01644, 2016.

\bibitem{ref3} Guillaume Astruc, Nicolas Gonthier, Clement Mallet, and Loic Landrieu. OmniSat: Self-supervised modality fusion for Earth observation. In European Conference on Computer Vision (ECCV), pages 409--427, 2024.

\bibitem{ref4} Guillaume Astruc, Nicolas Gonthier, Clement Mallet, and Loic Landrieu. AnySat: One Earth observation model for many resolutions, scales, and modalities. In Proceedings of the IEEE/CVF Conference on Computer Vision and Pattern Recognition (CVPR), pages 19530--19540, 2025.

\bibitem{ref5} Sarah Brood, Iris Dumeur, Jeremy Anger, Aurelien de Truchis, Ewan Sean, Ibrahim Fayad, Alexandre d’Aspremont, and Philippe Ciais. Addressing geographical domain shift in tree species mapping via foundation models using satellite image time series. EGU General Assembly 2026, abstract EGU26-13546, 2026.

\bibitem{ref6} Mathilde Caron, Hugo Touvron, Ishan Misra, Herve Jegou, Julien Mairal, Piotr Bojanowski, and Armand Joulin. Emerging properties in self-supervised vision transformers. In Proceedings of the IEEE/CVF International Conference on Computer Vision (ICCV), pages 9630--9640, 2021.

\bibitem{ref7} Yezhen Cong, Samar Khanna, Chenlin Meng, Patrick Liu, Erik Rozi, Yutong He, Marshall Burke, David B. Lobell, and Stefano Ermon. SatMAE: Pre-training transformers for temporal and multi-spectral satellite imagery. In Advances in Neural Information Processing Systems (NeurIPS), pages 197--211, 2022.

\bibitem{ref8} Isaac Corley, Caleb Robinson, Rahul Dodhia, Juan M. Lavista Ferres, and Peyman Najafirad. Revisiting pre-trained remote sensing model benchmarks: Resizing and normalization matters. In Proceedings of the IEEE/CVF Conference on Computer Vision and Pattern Recognition Workshops (CVPRW), pages 3162--3172, 2024.

\bibitem{ref9} Alexey Dosovitskiy, Lucas Beyer, Alexander Kolesnikov, Dirk Weissenborn, Xiaohua Zhai, Thomas Unterthiner, Mostafa Dehghani, Matthias Minderer, Georg Heigold, Sylvain Gelly, Jakob Uszkoreit, and Neil Houlsby. An image is worth 16x16 words: Transformers for image recognition at scale. In International Conference on Learning Representations (ICLR), 2021. arXiv:2010.11929.

\bibitem{ref10} Fabian Ewald Fassnacht, Hooman Latifi, Krzysztof Stere´nczak, Aneta Modzelewska, Michael Lefsky, Lars T. Waser, Christoph Straub, and Aniruddha Ghosh. Review of studies on tree species classification from remotely sensed data. Remote Sensing of Environment, 186:64--87, 2016.

\bibitem{ref11} Charles Gaydon and Floryne Roche. PureForest: A largescale aerial lidar and aerial imagery dataset for tree species classification in monospecific forests. In Proceedings of the IEEE/CVF Winter Conference on Applications of Computer Vision (WACV), pages 5895--5904, 2025.

\bibitem{ref12} Kaiming He, Xiangyu Zhang, Shaoqing Ren, and Jian Sun. Deep residual learning for image recognition. In Proceedings of the IEEE Conference on Computer Vision and Pattern Recognition (CVPR), pages 770--778, 2016.

\bibitem{ref13} Takayuki Ishikawa, Carmelo Bonannella, Bas J. W. Lerink, and Marc Ru\ss{}wurm. Assessing the effectiveness of deep embeddings for tree species classification in the Dutch forest inventory. arXiv preprint arXiv:2508.18829, 2025.

\bibitem{ref14} Teja Kattenborn, Felix Schiefer, Julian Frey, Hannes Feilhauer, Miguel D. Mahecha, and Carsten F. Dormann. Spatially autocorrelated training and validation samples inflate performance assessment of convolutional neural networks. ISPRS Open Journal of Photogrammetry and Remote Sensing, 5:100018, 2022.

\bibitem{ref15} Pang Wei Koh, Shiori Sagawa, Henrik Marklund, Sang Michael Xie, Marvin Zhang, Akshay Balsubramani, et al. WILDS: A benchmark of in-the-wild distribution shifts. In Proceedings of the 38th International Conference on Machine Learning (ICML), pages 5637--5664, 2021.

\bibitem{ref16} Ananya Kumar, Aditi Raghunathan, Robbie Jones, Tengyu Ma, and Percy Liang. Fine-tuning can distort pretrained features and underperform out-of-distribution. In International Conference on Learning Representations (ICLR), 2022. arXiv:2202.10054.

\bibitem{ref17} Hans R. Kunsch. The jackknife and the bootstrap for general stationary observations. The Annals of Statistics, 17(3): 1217--1241, 1989.

\bibitem{ref18} Alexandre Lacoste, Nils Lehmann, Pau Rodriguez, Evan David Sherwin, Hannah Kerner, Bjorn Lutjens, Jeremy Andrew Irvin, David Dao, Hamed Alemohammad, Alexandre Drouin, et al. GEO-Bench: Toward foundation models for earth monitoring. In Advances in Neural Information Processing Systems (NeurIPS), Datasets and Benchmarks Track, pages 51080--51093, 2023.

\bibitem{ref19} Dong C. Liu and Jorge Nocedal. On the limited memory BFGS method for large scale optimization. Mathematical Programming, 45(1--3):503--528, 1989.

\bibitem{ref20} Hong Liu, Jeff Z. HaoChen, Adrien Gaidon, and Tengyu Ma. Self-supervised learning is more robust to dataset imbalance. In International Conference on Learning Representations (ICLR), 2022. arXiv:2110.05025.

\bibitem{ref21} Valerio Marsocci, Yuru Jia, Georges Le Bellier, David Kerekes, et al. PANGAEA: Assessing geospatial foundation models capabilities through a global and inclusive benchmark. IEEE Geoscience and Remote Sensing Magazine, 14 (1):245--285, 2026.

\bibitem{ref22} Hanna Meyer and Edzer Pebesma. Machine learning-based global maps of ecological variables and the challenge of assessing them. Nature Communications, 13:2208, 2022.

\bibitem{ref23} Maxime Oquab, Timothee Darcet, Theo Moutakanni, Huy V. Vo, Marc Szafraniec, Vasil Khalidov, Pierre Fernandez, Daniel Haziza, Francisco Massa, Alaaeldin El-Nouby, et al. DINOv2: Learning robust visual features without supervision. Transactions on Machine Learning Research, 2024. arXiv:2304.07193.

\bibitem{ref24} Pierre Ploton, Frederic Mortier, Maxime Rejou-Mechain, Nicolas Barbier, Nicolas Picard, Vivien Rossi, Carsten Dormann, Guillaume Cornu, Ga\"{e}lle Viennois, Nicolas Bayol, et al. Spatial validation reveals poor predictive performance of large-scale ecological mapping models. Nature Communications, 11:4540, 2020.

\bibitem{ref25} David R. Roberts, Volker Bahn, Simone Ciuti, Mark S. Boyce, Jane Elith, Gurutzeta Guillera-Arroita, Severin Hauenstein, Jose J. Lahoz-Monfort, Boris Schroder, Wilfried Thuiller, David I. Warton, Brendan A. Wintle, Florian Hartig, and Carsten F. Dormann. Cross-validation strategies for data with temporal, spatial, hierarchical, or phylogenetic structure. Ecography, 40(8):913--929, 2017.

\bibitem{ref26} Felix Schiefer, Teja Kattenborn, Annett Frick, Julian Frey, Peter Schall, Barbara Koch, and Sebastian Schmidtlein. Mapping forest tree species in high resolution UAV-based RGB-imagery by means of convolutional neural networks. ISPRS Journal of Photogrammetry and Remote Sensing, 170:205--215, 2020.

\bibitem{ref27} Christian Schulz, Steve Ahlswede, Christiano Gava, Patrick Helber, Benjamin Bischke, Florencia Arias, Michael F{o}rster, J{o}rn Hees, Beg{u}m Demir, and Birgit Kleinschmit. TreeSatAI benchmark archive for deep learning in forest applications (v1.0.1). Zenodo, 2022.

\bibitem{ref28} Oriane Simeoni, Huy V. Vo, Maximilian Seitzer, Federico Baldassarre, Maxime Oquab, Cijo Jose, Vasil Khalidov, Marc Szafraniec, Seungeun Yi, Micha\"{e}l Ramamonjisoa, et al. DINOv3. arXiv preprint arXiv:2508.10104, 2025.

\bibitem{ref29} Jamie Tolan, Hung-I Yang, Benjamin Nosarzewski, Guillaume Couairon, Huy V. Vo, John Brandt, Justine Spore, Sayantan Majumdar, Daniel Haziza, Janaki Vamaraju, et al. Very high resolution canopy height maps from RGB imagery using self-supervised vision transformer and convolutional decoder trained on aerial lidar. Remote Sensing of Environment, 300:113888, 2024.

\bibitem{ref30} Roozbeh Valavi, Jane Elith, Jose J. Lahoz-Monfort, and Gurutzeta Guillera-Arroita. blockCV: An R package for generating spatially or environmentally separated folds for k-fold cross-validation of species distribution models. Methods in Ecology and Evolution, 10(2):225--232, 2019.

\end{thebibliography}
\end{document}